\documentclass[final]{cvpr}

\usepackage{times}
\usepackage{epsfig}
\usepackage{graphicx}
\usepackage{amsmath}
\usepackage{amssymb}
\usepackage{multirow}
\usepackage{algorithm}
\usepackage{algpseudocode}
\usepackage{amsmath}

\usepackage[pagebackref=false,breaklinks=true,colorlinks,bookmarks=false]{hyperref}

\def\cvprPaperID{1342} 
\def\confYear{CVPR 2021}

\begin{document}

\title{Diverse Part Discovery: Occluded Person Re-identification with \\Part-Aware Transformer}

\author{ Yulin Li$^1$\thanks{Equal contribution}~, Jianfeng He$^{1*}$, Tianzhu Zhang$^{1 }$\thanks{Corresponding author}\;, Xiang Liu$^2$, Yongdong Zhang$^1$, Feng Wu$^1$\\
$^1$ University of Science and Technology of China~~~~$^2$ Dongguan University of Technology\\
{\tt\small \{liyulin, hejf\}@mail.ustc.edu.cn~~~\{tzzhang, fengwu, zhyd73\}@ustc.edu.cn}
\\{\tt\small succeedpkmba2011@163.com}
}

\maketitle
\pagestyle{empty}  
\thispagestyle{empty} 

\begin{abstract}
Occluded person re-identification (Re-ID) is a challenging task 
as persons are frequently occluded by various obstacles or other persons, 
especially in the crowd scenario.
To address  these issues,
we propose a novel end-to-end Part-Aware Transformer (PAT)
for occluded person Re-ID
through diverse part discovery
via a transformer encoder-decoder architecture, including a
pixel context based transformer encoder and
a part prototype based transformer decoder.
The proposed PAT model enjoys several merits.
First,  to the best of our knowledge, 
this is the first work to exploit  
the transformer encoder-decoder architecture 
for occluded person Re-ID in a unified deep model.
Second,  to learn part prototypes well  with only identity labels, 
we design two effective mechanisms  
including part diversity   and part discriminability.
Consequently, we can achieve diverse part discovery for 
occluded person Re-ID in a weakly supervised manner.
Extensive experimental  results on six challenging benchmarks
for three tasks (occluded, partial and holistic Re-ID)
demonstrate that 
our proposed PAT performs favorably against   stat-of-the-art methods.

\end{abstract}

\vspace{-2mm}
\section{Introduction}	
\vspace{-1mm}
Person re-identification (Re-ID) aims to match images of a person captured from non-overlapping camera views~\cite{gong2011person,xiong2014person,zheng2016person}.
%
It is one of the most  important research topics in the computer vision field with various applications, such as  video surveillance,  autonomous driving, and activity analysis~\cite{yang2014salient,liao2015person,zheng2012reidentification,RSST_TPAMI19,MCPF_TPAMI19}.
Recently, person Re-ID has drawn a growing amount of interest from academia and industry,
and various methods have been proposed~\cite{koestinger2012large,liao2015efficient,hermans2017defense,sun2018beyond,jiang2020self,lu2020cross,wang2019rgb,wang2020cross}.
Most of these approaches assume that the entire body of the pedestrian is available for
the Re-ID model designing.
However, when conducting person Re-ID in real-world scenarios, e.g., airports, railway stations, hospitals, and malls, it is difficult to satisfy this
assumption due to the  inevitable occlusions.
For example, a person may be occluded by some
obstacles (e.g., cars, trees,   walls,  and other persons), and the camera fails to capture the holistic person.
%
%
%
Therefore, it is essential to design an effective model to solve this occluded
person Re-ID problem~\cite{zhuo2018occluded,miao2019pose}.

%


\begin{figure}[t]
\includegraphics[width=1.0\linewidth]{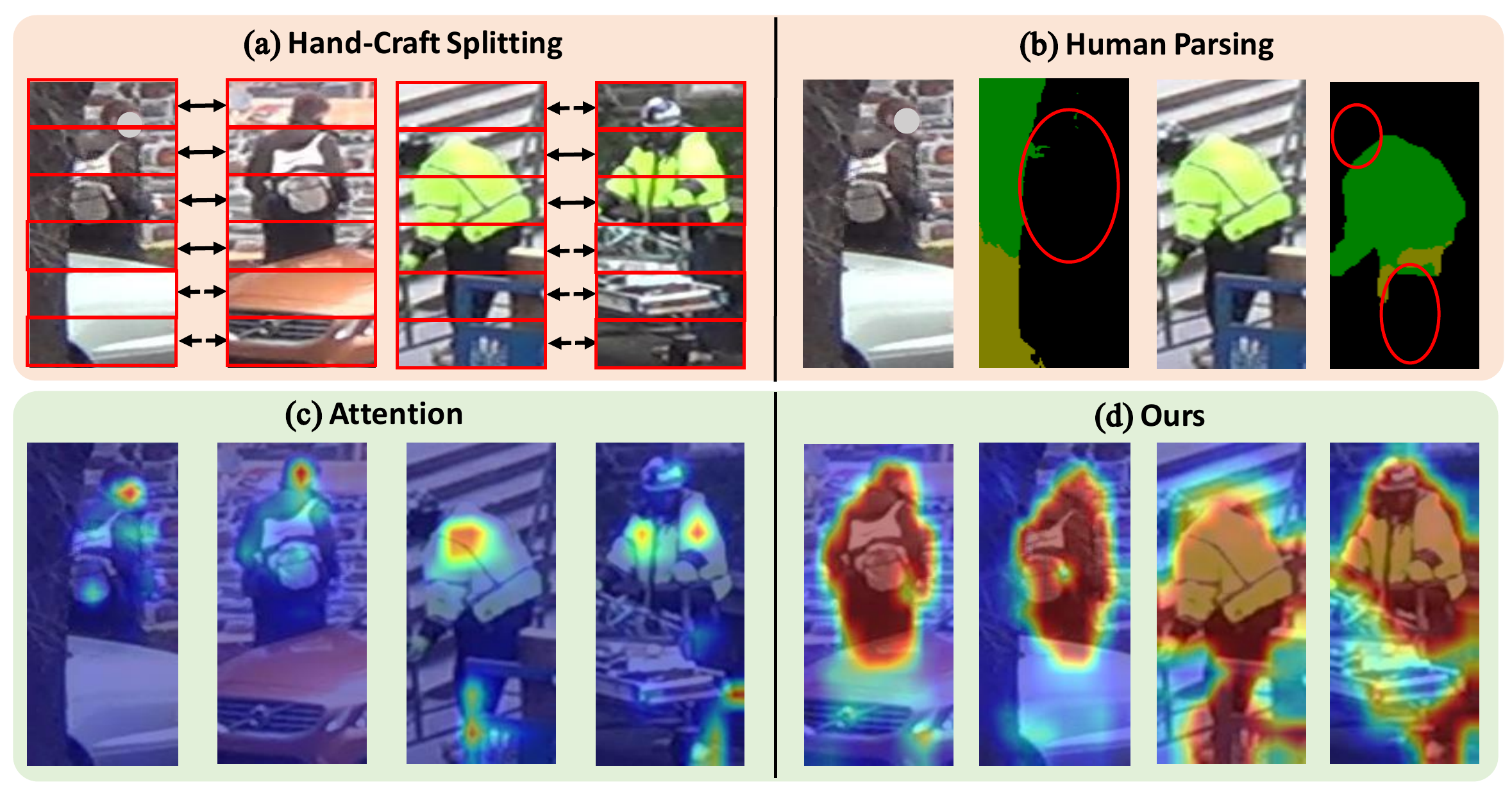}
\caption{
Examples of part-based methods for occluded person Re-ID.
(a) The hand-crafted splitting based methods  require  strict person alignment.
(b) The extra semantic based methods exclude  the personal belongings and are error-prone
when a person is seriously occluded.
(c) The attention-based methods tend  to mainly focus on the most discriminative region.
(d) The attention maps produced by our proposed PAT by fusing all part-aware masks.
The discriminative parts are highlighted.
        }
\label{fig1}
\vspace{-5mm}
\end{figure}

In the occluded person Re-ID task, occluded regions usually contain some noise that results in mismatching, and the key issue is how to learn discriminative features from unoccluded regions.
%
%
Recently, leveraging local features extracted from human body parts to improve representations of the pedestrian has been the mainstream for robust feature learning of the occluded Re-ID task.
Generally, these part-based occluded Re-ID methods can be divided into three main categories.
(1) The hand-crafted splitting based methods divide the image or feature map into small patches~\cite{he2018deep,zheng2015partial,he2018recognizing} or rigid stripes~\cite{sun2018beyond,fan2018scpnet} and then extract part features from the local patches or stripes.
However, hand-crafted splitting is too coarse to align the human parts well and introduces lots of background noise.
%
(2) The extra semantic based methods~\cite{he2019foreground,miao2019pose,wang2020high,gao2020pose,he2020guided} directly utilize human parsing~\cite{he2019foreground,he2020guided} or pose estimation models~\cite{miao2019pose,wang2020high,gao2020pose} as part localization modules to achieve more accurate human part localization.
%
However, their success heavily relies on the accuracy of the off-the-shelf human parsing or pose estimation models.
Since there exist differences between training datasets of human parsing/pose estimation and person Re-ID, the off-the-shelf human parsing/pose estimation models are error-prone when pedestrians are seriously occluded.
(3) The attention based methods~\cite{sun2019perceive,zhuo2019novel} exploit attention mechanisms to localize discriminative human parts.
Typically, the predicted attention maps distribute most of the attention weights on human parts, which can help decrease the negative effect of cluttered background.
%
To sum up, most existing occluded Re-ID methods focus on locating discriminative human parts and leveraging local part features to develop powerful representations of the pedestrian.

Based on the above discussions,
the part-based representations have been proven to be effective for the occluded Re-ID problem.
To capture accurate human parts, an intuitive idea is to detect non-occluded body parts using body part detectors and then match the corresponding body parts.
However, there are no extra annotations for the body detector learning.
Thus, we propose to localize   discriminative human parts only with identity labels.
To achieve this goal, there are two main challenges as follows.
%
On the one hand,  background with diverse characteristics, such as colors, sizes, shapes, and positions, increase the difficulty of getting robust features for the target person.
Intuitively, the appearance of pixels of the same human part region is similar, while quite different from the background pixels.
Therefore, it is necessary to model the correlation between pixels for robust feature representation.
%
%
On the other hand, as shown in Figure~\ref{fig1}, the occluded parts vary between different pedestrian images.
As there are no groundtruth annotations for  human parts, it is difficult to cope with diverse appearance of pedestrians and adaptively locate all unoccluded parts only with the identity labels.
%
%
%
As a result, as shown in Figure~\ref{fig1} (c),
most of the attention based methods   tend  to put the main focus on the most discriminative region.
%
They always ignore other human parts including personal belongings, e.g., backpack and reticule,
which also provide important clues  for person Re-ID.
%

To deal with the above issues,
we propose a novel Part-Aware Transformer (PAT)
for occluded person Re-ID through
diverse part discovery via a transformer encoder-decoder architecture~\cite{vaswani2017attention,carion2020end},
including a pixel context based transformer encoder
and a part prototype based transformer decoder.
%
%
%
%
In the \textbf{pixel context based transformer encoder},
we adopt a self-attention mechanism to
capture the full image context information.
Specifically, we model the correlation of
pixels of the feature map and
aggregate pixels with similar appearances.
In this way, we can obtain the pixel context aware
feature map, which is more robust to background clutters.
%
%
In the \textbf{part prototype based transformer decoder},
we introduce a set of learnable part prototypes
to generate part-aware masks focusing on discriminative human parts.
%
%
In specific, given the feature map of a pedestrian,
we take the learnable part prototypes
as queries and pixels of the feature map as keys and
values of the transformer decoder.
We can obtain part-aware masks by
calculating the similarity between all pixels
in the feature map and part prototypes.
Each part-aware mask is expected to
denote the spatial distribution of
one specific human part, e.g., head or body part.
With part-aware masks, human part features
can be further obtained from the values
by a weighted pooling.
%
%
%
However, without the assistance of part annotations,
it is challenging to constraint these part prototypes to
capture accurate human parts.
Thus, to guide   part prototype learning,
we propose two mechanisms
including part diversity
and part discriminability.
Intuitively, different part features of the same pedestrian
should focus on different human parts.
Therefore, the part diversity mechanism
is adopted to encourage lower correlation
between part features and make part prototypes
focus on different discriminative foreground regions.
The part discriminability mechanism is to make part features maintain identity discriminative via  part classification and a triplet loss.
%
%
By optimizing the transformer encoder and decoder jointly,   part prototypes can be learned through the whole dataset.
Consequently, we can achieve robust human part discovery for   occluded person Re-ID   in a weakly supervised manner.

The contributions of our method could be summarized into three-fold:
(1) We propose a novel end-to-end Part-Aware Transformer
for occluded person Re-ID
through diverse part discovery
via a transformer encoder-decoder architecture, including a
pixel context based transformer encoder and
a part prototype based transformer decoder.
%
%
To the best of our knowledge, our PAT is the first work by exploiting the transformer encoder-decoder architecture for occluded person Re-ID in a unified deep model.
(2) To learn part prototypes only with identity labels well, we design two effective mechanisms, including part diversity   and part discriminability.
Consequently, we can achieve robust human part discovery for occluded person Re-ID in a weakly supervised manner.
%
%
(3) To demonstrate the effectiveness of our method, we perform experiments on
three tasks, including occluded Re-ID, partial Re-ID and holistic Re-ID
on six standard Re-ID datasets.
Extensive experimental results demonstrate that the proposed method performs favorably against state-of-the-art methods.

\vspace{-1.5mm}
\section{Related Work}
\vspace{-1mm}
In this section, we briefly overview methods
that are related to holistic person Re-ID,
partial Re-ID and  occluded person Re-ID respectively.

\noindent\textbf{Holistic Person Re-Identification.} 
%
Person re-identification (Re-ID) aims to match images of a person captured from non-overlapping camera views~\cite{gong2011person,xiong2014person,zheng2016person}.
Existing Re-ID methods can be summarized to hand-crafted descriptors~\cite{yang2014salient,liao2015person}, metric learning methods~\cite{zheng2012reidentification,koestinger2012large,liao2015efficient} and deep learning methods\cite{sun2018beyond,liu2018pose,saquib2018pose,su2017pose,xu2018attention,zheng2019pose,kalayeh2018human,li2018harmonious,song2018mask,liu2017hydraplus,li2018unified}.
Recent works utilizing part-based features have achieved state-of-the-art performance for the holistic person Re-ID task.
Kalayeh \etal~\cite{kalayeh2018human} extract several region parts with human parsing methods and assemble final discriminative representations with part-level features.
Sun \etal~\cite{sun2018beyond} uniformly partition the feature map and learn part-level features by multiple classifiers.
Zhao \etal~\cite{zhao2017deeply} and Liu \etal~\cite{liu2017hydraplus} extract part-level features by attention-based methods.
But all these Re-ID methods focus  on matching holistic person images with the assumption that the entire body of the pedestrian is available.
%
Different from these methods, our model can adaptively capture discriminative human part features via a transformer encoder-decoder architecture for the occluded person Re-ID task.

%

\noindent\textbf{Partial Person Re-Identification.}
Partial person Re-ID aims to match partial probe images to holistic gallery images.
Zheng \etal~\cite{zheng2015partial} propose a local-level matching model called Ambiguity-sensitive Matching Classifier (AMC) based on the dictionary learning and introduce a local-to-global matching model called Sliding Window Matching 
to provide complementary spatial layout information.
%
He \etal~\cite{he2018deep} propose
an alignment-free approach namely Deep Spatial feature
Reconsruction (DSR) that exploits the reconstruction error
based on sparse coding.
%
Luo et al.\etal~\cite{luo2020stnreid} proposed STNReID that combines
a spatial transformer network (STN) and a Re-ID network for partial Re-ID.
Sun \etal~\cite{sun2019perceive}
introduce a Visibility-aware Part Model (VPM)
to perceive the visibility of part regions through self-supervision.
However, all these methods need a
manual crop of the occluded target person
in the probe image and then use the non-occluded
parts as the new query.
The manual cropping is not efficient in practice and
 might introduce human bias to
the cropped results.

\noindent\textbf{Occluded Person Re-Identification.}
Given occluded probe images, occluded person Re-ID aims to find the same person with holistic or occluded appearance in disjoint cameras.
This task is more challenging due to incomplete
information and spatial misalignment.
%
Zhuo \etal~\cite{zhuo2018occluded} combine
the occluded/unoccluded classification task
and person ID classification task to
extract key information from images.
%
%
He \etal~\cite{he2019foreground} reconstruct
the feature map of unoccluded regions and propose
a spatial background-foreground classifier 
to avoid the influence of background clutters.
%
%
Besides, the Pose-Guided Feature Aligment (FGFA)~\cite{miao2019pose} utilizes   pose landmarks to mine discriminative parts to address the occlusion noise.
Gao \etal~\cite{gao2020pose} propose a Pose-guided Visible Part Matching (PVPM) model to learn discriminative part features with pose-guided attentions.
%
%
Wang \etal~\cite{wang2020high} exploit graph convolutional layers to learn high-order human part relations   for   robust alignment.
Although the above methods can solve the occlusion problem to some extent, most of them heavily rely on off-the-shelf human parsing models or pose estimators.
%
Different from them,
our model can exploit diverse parts     with  only identity labels
in a weakly supervised manner via a transformer
encoder-decoder architecture.



\vspace{-1.5mm}
\section{Part-Aware Transformer}
\vspace{-1mm}
\begin{figure*}
\centering
\includegraphics[width=0.95\linewidth]{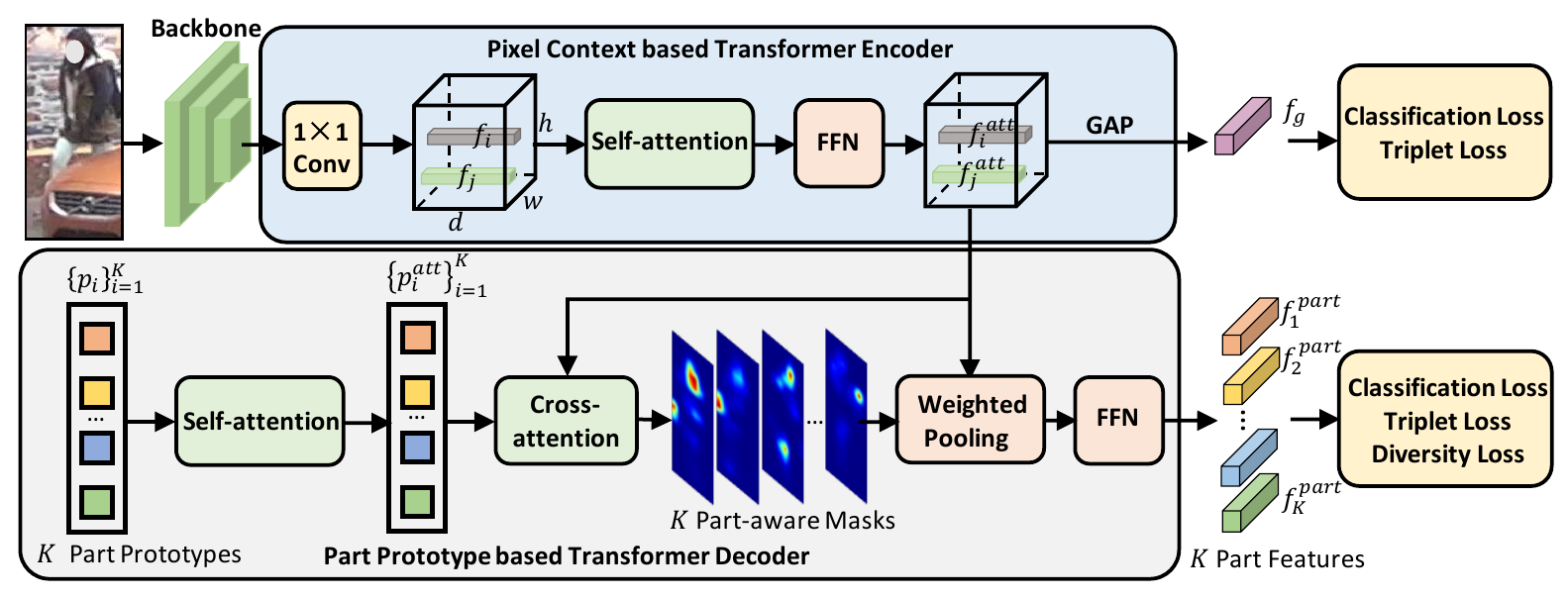}
\caption{
The pipeline of the proposed PAT consists of 
a pixel context based transformer encoder 
and a part prototype based transformer decoder.
%
Here, ``self-attention" denotes the self-attention layer,
``cross-attention" denotes the cross-attention layer,
and ``FFN" denotes the feed forward layer.
For more details, please refer to the text and please see supplemental materials for detailed architecture.
}\label{fig2}
\vspace{-5mm}
\end{figure*}

In this section, we introduce the proposed Part-Aware Transformer (PAT) in detail.
As shown in Figure~\ref{fig2}, the proposed PAT mainly consists of two modules, including
the pixel context based transformer encoder
and the part prototype based transformer decoder.
Here we give a brief introduction to the full process.
First, we obtain the feature map of each pedestrian image through a CNN backbone.
%
Then we flatten the feature map and carry
out the self-attention operation to obtain the pixel context aware feature map
with the transformer encoder.
After obtaining the pixel context aware
feature map, we calculate the similarity
between the feature map and a set of learnable part
prototypes to obtain part-aware masks.
Part features can be further obtained
by a weighted pooling
where part-aware masks are
treated as different spatial attention maps.
Finally, we introduce the part diversity mechanism
and part discriminability mechanism to learn part prototypes well with only identity labels.

\subsection{Pixel Context based Transformer Encoder}\label{subsec:pexelcontext}
\vspace{-1mm}
%
Background regions with diverse characteristics increase the difficulty of getting robust features for the target person.
Therefore, we adopt a self-attention mechanism to capture the full image
context information.
In this way, we can obtain the pixel
context aware feature map, which is more robust to background
clutters.
Following~\cite{sun2018beyond}, our method uses ResNet-50~\cite{he2016deep} without the average pooling layer and fully connected layer as the backbone to extract global feature maps from given images.
We also set the stride of conv4\_1 to 1 to increase the feature resolution as in~\cite{sun2018beyond}.
As a result, an input image with a size of $H \times W$
will get the feature map with
the spatial dimension of $H/16 \times W/16$,
which is larger than that of the original ResNet-50.
A larger feature map
has been proved to be effective
in person Re-ID.
%
Formally, the feature map extracted from
the backbone is denoted as
$\mathbf{Z} \in \mathbb{R}^{h \times w \times c}$,
where $h,w,c$ are the height, width and channel of the global feature map, respectively.
We first utilize a $1 \times 1$ convolution
to reduce the channel dimension of
the feature map $\mathbf{Z}$ to
a smaller dimension $d$,
creating a new feature map
$\mathbf{F} \in \mathbb{R}^{h \times w \times d}$.
The transformer encoder requires a 1D sequence as input.
To handle 2D feature maps,
we flatten the spatial dimensions of $\mathbf{F}$
into one dimension, resulting in a $hw \times d$ feature.
%
%
%
In the self-attention mechanism,
given the feature map
$\mathbf{F}=\left[f_{1} ; f_{2} ; \ldots ; f_{hw}\right]$ ($f_{i} \in \mathbb{R}^{1 \times d}$ indicates the feature of the $i^{th}$ spatial position),
both keys, queries and values arise from pixels of the feature map.
Formally,
\begin{equation}
{\mathbf{Q}}_{i} = f_{i} \mathbf{W}^{Q}, \quad {\mathbf{K}}_{j} = f_{j} \mathbf{W}^{K}, \quad {\mathbf{V}}_{j} = f_{j} \mathbf{W}^{V},
\end{equation}
where $i, j \in {1,2, \ldots , hw}$ and $\mathbf{W}^{Q} \in \mathbb{R}^{d \times d_k}, \mathbf{W}^{K} \in \mathbb{R}^{d \times d_k}, \mathbf{W}^{V} \in \mathbb{R}^{d \times d_v}$ are linear projections.
%
%
For the $i^{th}$ query ${\mathbf{Q}}_{i}$,  the attention weights are calculated based on the dot-product similarity between each query and the keys:
\begin{equation}\label{eq:attention}
s_{i, j}=\frac{\exp \left(\beta_{i, j}\right)}{\sum_{j=1}^{hw} \exp \left(\beta_{i, j}\right)},
\beta_{i, j}=\frac{{\mathbf{Q}}_{i} {\mathbf{K}}_{j}^{T}}{\sqrt{d_{k}}},
\end{equation}
where $\sqrt{d_{k}}$ is a scaling factor.
The output of the self-attention mechanism is defined as weighted sum over all values according to the attention weights:
\begin{equation}\label{eq:attention}
\hat{f}^{att}_{i}= \operatorname{Att}\left({\mathbf{Q}}_{i},\mathbf{K}, \mathbf{V} \right) =\sum_{j=1}^{hw} s_{i, j} {\mathbf{V}}_{j},
\end{equation}
The normalized attention weight $s_{i, j}$ models   the interdependency between different spatial pixels $f_{i}$ and $f_{j}$,
and the weight sum of the values
can aggregate these semantically related
spatial pixels to update $f_{i}$.
Since pixels of the same human part   have high similarities while are different from background pixels, the feature map capturing the full-image context information would be more robust to background clutters.
We implement Eq.(\ref{eq:attention}) with the multi-head attention mechanism and can get $\hat{f}^{att}_{i} \in \mathbb{R}^{1 \times d}$.
The updated feature map can be
obtained by aggregating
pixel context information of all positions:
%
\begin{equation}\label{eq:hat}
{\mathbf{\hat{F}}}^{att} = \left[\hat{f}^{att}_{1} ; \hat{f}^{att}_{2}; \ldots ; \hat{f}^{att}_{hw}\right] \in \mathbb{R}^{hw \times d},
\end{equation}
where ${\mathbf{\hat{F}}}^{att}$ represents
the updated feature map with the self-attention mechanism.
Following the standard transformer architecture, we use the feed-forward network to produce the final pixel context aware feature map as defined in:
\begin{equation}\label{eq:FFN}
{\mathbf{F}}^{att} =\operatorname{FFN}({\mathbf{\hat{F}}}^{att}),
\end{equation}
%
%
%
where $\operatorname{FFN}(\cdot)$ is a simple neural network
using two fully connected layers~\cite{vaswani2017attention}.
The residual connections followed by the layer normalization~\cite{ba2016layer} are also applied.
Please see supplemental materials for more details about the architecture.
Through the self-attention operation,
the pixel context aware feature map
${\mathbf{F}}^{att} = \left[{f}^{att}_{1} ; {f}^{att}_{2}; \ldots ; {f}^{att}_{hw}\right] \in \mathbb{R}^{hw \times d}$
can be obtained, which is more robust to background clutter.

\noindent\textbf{Encoder Training Loss.}
To make the pixel context aware feature focus on ID-related discriminative information and train our encoder, we use an identity classification loss and a triplet loss as the objective function.
First of all, we utilize a global average pooling operation $f^{g} = GAP({\mathbf{F}}^{att})$ and constrain the global feature $f^{g} \in \mathbb{R}^{1 \times d}$ to satisfy the objective function.
The objective function   is   formulated as:
\begin{equation}\label{eq:encoderloss}
\begin{aligned}
\mathcal{L}_{En} &={\lambda}_{cls}\mathcal{L}_{c l s}\left(f^{g}\right)+ {\lambda}_{tri}\mathcal{L}_{t r i}\left(f^{g}\right) \\
& =-{\lambda}_{cls}\log p_{g} + {\lambda}_{tri}\left[\alpha+d_{f^{g}, f_{p}^{g}}-d_{f^{g}, f_{n}^{g}}\right]_{+}.
\end{aligned}
\end{equation}
Where $p_{g}$ is the probability predicted by a classifier,
$\alpha$ is the margin, $d_{f^{g}, f_{p}^{g}}$ is the distance between a positive pair $\left(f^{g}, f_{p}^{g}\right)$ from the same identity, and $\left(f^{g}, f_{n}^{g}\right)$ is the negative pair from different identities.
%

\subsection{Part Prototype based Transformer Decoder}
\vspace{-1mm}
%
%
In the part prototype based transformer decoder, to localize discriminative human parts only with identity labels, we introduce a set of learnable part prototypes focusing on discriminative human parts and propose two mechanisms, including part diversity and part discriminability to guide part prototype learning only with identity labes.
In this way, we can achieve robust human part discovery in a weakly supervised manner.
First of all, we introduce a set of part prototypes $\mathcal{P}_{K}=\left\{p_{i}\right\}_{i=1}^{K}, p_i \in \mathbb{R}^{1 \times d}$
represents a part classifier that
determines whether pixels of the feature
map ${\mathbf{F}}^{att}$ belong  to the part $i$.
These part prototypes are set as learnable parameters.

\noindent\textbf{Self-attention Layer.}
Following the standard architecture of the transformer,
we first use a self-attention mechanism
to further incorporate the local context of
human parts to part prototypes.
%
This process allows the local context information
propagation between prototypes during part prototype learning.
The implementation is the same as in Section~\ref{subsec:pexelcontext}, and both keys,
queries and values arise from part prototypes.
We can obtain the updated part prototype set $\left\{p_{i}^{att}\right\}_{i=1}^{K}$.
%
%
The weights of self-attention
encode the relations between
part prototypes $p_{i}$ and $p_{j}$.
The updated part prototypes
incorporate the local context of different parts.

\noindent\textbf{Cross-attention Layer.}
The cross-attention layer aims to
extract foreground part features
from the feature map ${\mathbf{F}}^{att}$ with the learnable part prototypes.
As shown in Figure~\ref{fig2},
in the cross-attention layer,
given the feature map
${\mathbf{F}}^{att} = \left[f^{att}_{1} ; f^{att}_{2} ; \ldots ; f^{att}_{hw}\right]$,
queries arise from part prototypes
$\left\{p_{i}^{att}\right\}_{i=1}^{K}$,
keys and values arise from pixels of the feature map.
Formally,
\begin{equation}
{\mathbf{Q}}_{i} = p_{i}^{att} \mathbf{W}^{Q}, \; {\mathbf{K}}_{j} = f^{att}_{j} \mathbf{W}^{K}, \; {\mathbf{V}}_{j} = f^{att}_{j} \mathbf{W}^{V},
\end{equation}
where $i \in 1, 2, \dots, K$, $j \in 1, 2, \dots, hw$,  and
$\mathbf{W}^{Q} \in \mathbb{R}^{d \times d_k}, \mathbf{W}^{K} \in \mathbb{R}^{d \times d_k}, \mathbf{W}^{V} \in \mathbb{R}^{d \times d_v}$ are linear projections.
Note that they are different from Eq.(\ref{eq:attention}).
For each part prototype $p^{att}_{i}$,
we illustrate how to compute
the part-aware mask and the corresponding part feature. Formally,
\begin{equation}\label{eq:part-mask}
m_{i, j}=\frac{\exp \left(\beta_{i, j}\right)}{\sum_{j=1}^{hw} \exp \left(\beta_{i, j}\right)},
\beta_{i, j}=\frac{{\mathbf{Q}}_{i} {\mathbf{K}}_{j}^{T}}{\sqrt{d_{k}}},
\end{equation}
where $\sqrt{d_{k}}$ is a scaling factor.
The attention weight $m_{i, j}$ indicates the probability of the spatial feature $f^{att}_{j}$ belonging to the foreground part $i$.
The attention weights of all $hw$
positions make up a part-aware mask ${\mathbf{M}}_{i} =\left[m_{i, 1} ; m_{i, 2} ; \ldots ; m_{i, hw}\right]$, which has high response values at pixels belonging to the part $i$.
We can further obtain $i^{th}$ part feature
by a weighted pooling,
which is defined as the weighted sum over all values:
\begin{equation}\label{eq:part-feature}
\hat{f}^{part}_{i}= \operatorname{Att}\left({\mathbf{Q}}_{i},\mathbf{K}, \mathbf{V} \right) = \sum_{j=1}^{hw} m_{i, j} {\mathbf{V}}_{j},
\end{equation}
%
%
By computing over all part prototypes,
we can obtain $K$ part-aware masks
(each mask is a $h \times w$ attention map)
and further obtain $K$ part features,
as shown in Figure~\ref{fig2}.
We implement Eq.(\ref{eq:part-feature}) with the multi-head attention mechanism and can get $\hat{f}^{part}_{i} \in \mathbb{R}^{1 \times d}$.
Then, two fully-connected layers
are adopted, which is the same as the standard transformer architecture.
%
The final part feature is formulated as:
\begin{equation}\label{eq:final-part}
f_i^{part} = \operatorname{FFN}(\hat{f}^{part}_{i}),
\end{equation}
where $i \in 1, 2, \dots, K$
and $\operatorname{FFN}(\cdot)$ denotes the feed-forward network as in Eq.(\ref{eq:FFN}).

%

Since there are no human part annotations,
part prototype learning tends to focus on
the same discriminative part (e.g., the body),
which may result  in a suboptimal solution.
Thus, to learn part prototypes only with identity labes,
we propose two mechanisms  including part diversity
and part discriminability.
(1) The part diversity mechanism is  to
make   part prototypes focus on
different discriminative foreground parts.
%
A diversity loss is imposed to expand the discrepancy among different part features ${\left\{ f_i^{part} \right\}}_{i=1}^{K}$:
\begin{equation}\label{eq:diversityloss}
\mathcal{L}_{div}=\frac{1}{K(K-1)} \sum_{i=1,j=1}^{K} \sum_{ i \neq j}^{K} \frac{\left\langle f_i^{part}, f_j^{part}\right\rangle}{\left\|f_i^{part}\right\|_{2}\left\|f_j^{part}\right\|_{2}},
\end{equation}
The intuition behind this loss is obvious.
If the $i^{th}$ and the $j^{th}$ prototypes give a high attention weight to the same foreground part, the $\mathcal{L}_{div}$ will be large and prompt these prototypes to adjust themselves adaptively.
%
%
%
(2) The part discriminability
mechanism is to make part features maintain
identity discriminative.
%
The part classification and triplet loss are employed to guide   part feature representation learning as in Eq.(\ref{eq:discriminloss}), where the definitions of $\mathcal{L}_{cls}(\cdot)$ and $\mathcal{L}_{t r i}(\cdot)$ can be found in Eq.(\ref{eq:encoderloss}).
\begin{equation}\label{eq:discriminloss}
\mathcal{L}_{dis} ={\lambda}_{cls} \sum_{i=1}^{K}\mathcal{L}_{c l s}\left(f_i^{part}\right)+ {\lambda}_{tri} \sum_{i=1}^{K}\mathcal{L}_{tri}\left(f_i^{part}\right).
\end{equation}
In the triplet loss, part features $f_i^{part}$ from different identities form negative pairs, and those from the same identity form positive pairs.
As a result, the features obtained from the same prototype with different identities are pushed away and the identity discriminative part features can be obtained.

%
%

\subsection{Training and Inference}
\vspace{-1mm}

For the occluded person Re-ID task,  our proposed PAT
is trained by minimizing the overall
objective with identity labels  as defined in Eq.(\ref{eq:overallloss}).
\begin{equation}\label{eq:overallloss}
\mathcal{L}_{PAT} = \mathcal{L}_{En} + \mathcal{L}_{div} + \mathcal{L}_{dis},
\end{equation}
During the testing stage, for each image of an unseen identity,
%
we concatenate the global feature $f^{g}$
and part features ${\left\{ f_i^{part} \right\}}_{i=1}^{K}$
as its representation:
\begin{equation}\label{eq:concat}
v=\left[f^{g}, f_1^{part}, \cdots, f_K^{part}\right].
\end{equation}
where $[\cdot]$ denotes a concatenation operation.
%

\vspace{-1.5mm}
\section{Experiments}
\vspace{-1mm}
%
%
In this section, we first verify the effectiveness of our proposed model for occluded person Re-ID, partial Re-ID, and  holistic Re-ID.
Then, we report a set of ablation studies to validate the effectiveness of each component.
Finally, we provide more visualization results.
%

\vspace{-1.2mm}
\subsection{Datastes and Evaluation Metrics}
\vspace{-1mm}
%
%
To demonstrate the effectiveness of our method,
we  conduct extensive experiments
on two occluded datasets: Occluded-Duke \cite{miao2019pose} and Occluded REID \cite{miao2019pose},
two partial Re-ID datasets: Partial-REID \cite{zheng2015partial} and Partial-iLIDS \cite{zheng2011person},
and two holistic Re-ID datasets: Market-1501 \cite{zheng2015scalable} and DukeMTMC-reID \cite{ristani2016performance,zheng2017unlabeled}.
The details are as follows.

\noindent\textbf{Occluded-Duke}~\cite{miao2019pose}
contains 15,618 training images, 17,661 gallery images, and 2,210 occulded query images.
It is selected from DukeMTMC-reID by leaving occluded images and filtering out some overlap images.
%

\noindent\textbf{Occluded-REID}~\cite{zhuo2018occluded}
is an occluded person dataset captured by mobile cameras, including 2,000 images belonging to 200 identities.
Each identity has five full-body person images and five occluded person images with different
viewpoints and different types of severe occlusions.

\noindent\textbf{Partial-REID}~\cite{zheng2015partial}
is a specially designed partial person Re-ID benchmark
that includes 600 images from 60 people, with five full-body images in gallery set and five partial images in query set per person.

\noindent\textbf{Partial-iLIDS}~\cite{he2018deep}
is a partial person Re-ID dataset based on the iLIDS dataset~\cite{zheng2011person},
and contains a total of 238 images from 119 people captured by multiple cameras in the airport,
and their occluded regions are manually cropped.
%

\noindent\textbf{Market-1501}~\cite{zheng2015scalable}
consists of 1,501 identities captured by 6 cameras.
The training set consists of 12,936 images of 751 identities, the query set consists of 3,368 images, and the gallery set consists of 19,732 images.

\noindent\textbf{DukeMTMC-reID}~\cite{ristani2016performance,zheng2017unlabeled}
contains 36,411 images of 1,404 identities captured by 8 cameras. The training set contains 16,522 images, the query set consists of 2,228 images and the gallery set consists of 17,661 images.

\noindent\textbf{Evaluation Metrics.}
We adpot standard metrics as in most person Re-ID literature, namely Cumulative Matching Characteristic (CMC) curves and mean average precision (mAP), to evaluate the quality of different Re-ID models.

\vspace{-1.2mm}
\subsection{Implementation Details}
\vspace{-1mm}
We adopt
ResNet-50~\cite{he2016deep}  pretrained on ImageNet
as our backbone by removing the global average pooling (GAP) layer and fully connected layer.
For classifiers, as in~\cite{luo2019bag} we use a batch normalization layer~\cite{ioffe2015batch} and a fully connected layer followed by a softmax function.
The number of part prototypes $K$ is set to 6 on Market-1501, and set to 14 on all other datasets.
The images are resized to 256 $\times$ 128 and augmented with random horizontal flipping, padding 10 pixels, random cropping, and random erasing~\cite{zhong2020random}.
Extra color jitter is adopted on occluded-REID and partial datasets to avoid domain variance.
The batch size is set to 64 with 4 images per person. During the training stage, all the modules are jointly trained  for 120 epochs.
The learning rate is initialized to $3.5 \times 10^{-4}$ and
decayed to its $0.1×$ and $0.01×$ at 40 and 70 epochs.
%

\vspace{-1mm}
\subsection{Comparison with State-of-the-art Methods}
\vspace{-1mm}

\noindent\textbf{Results on Occluded Re-ID Datasets.}
\begin{table}[]
\centering
\small
\caption{Performance comparison with state-of-the-arts on     Occluded-Duke and Occluded-REID. Our method achieves the best performance on the two occluded datasets.}
\label{tab:occulded_dataset}
\begin{tabular}{c|cc|cc}
\hline
\multirow{2}{*}{Methods} & \multicolumn{2}{c|}{Occluded-Duke} & \multicolumn{2}{c}{Occluded-REID} \\
\cline{2-5}
                    &Rank-1   &mAP    &Rank-1   &mAP    \\
\hline
\hline
Part-Aligned~\cite{zhao2017deeply}			&28.8		&20.2		&-		&-			\\
PCB~\cite{sun2018beyond}					&42.6		&33.7		&41.3	&38.9		\\
\hline
Part Bilinear~\cite{suh2018part}			&36.9		&-			&-		&-			\\
FD-GAN~\cite{ge2018fd}						&40.8		&-			&-		&-			\\
\hline
AMC+SWM~\cite{zheng2015partial}     &-      &-      &31.2   &27.3   \\
DSR~\cite{he2018deep}         &40.8   &30.4   &72.8   &62.8   \\
SFR~\cite{he2018recognizing}      &42.3   &32     &-      &-      \\
\hline
Ad-Occluded~\cite{huang2018adversarially}   &44.5   &32.2   &-      &-      \\
FPR~\cite{he2019foreground}       &-      &-      &78.3   &68.0   \\
PVPM~\cite{gao2020pose}         &47     &37.7   &70.4   &61.2   \\
PGFA~\cite{miao2019pose}        &51.4   &37.3   &-      &-      \\
GASM~\cite{he2020guided}        &-      &     &74.5   &65.6   \\
HOReID~\cite{wang2020high}        &55.1   &43.8   &80.3   &70.2   \\
ISP~\cite{zhu2020identity}        &62.8   &52.3   &-      &-      \\
\hline
\textbf{PAT}(\textit{Ours})          &\textbf{64.5}  &\textbf{53.6}  &\textbf{81.6}  &\textbf{72.1}  \\
\hline
\end{tabular}
\vspace{-4mm}
\end{table}
Table~\ref{tab:occulded_dataset} shows the performance of our model  and previous methods on two occluded datasets.
Four kinds of methods are compared, which are hand-crafted splitting based Re-ID methods~\cite{zhao2017deeply,sun2018beyond}, holistic Re-ID methds with key-point information~\cite{suh2018part,ge2018fd}, partial ReID methods~\cite{zheng2015partial,he2018deep,he2018recognizing} and occluded ReID methods~\cite{he2019foreground,gao2020pose,miao2019pose,he2020guided,wang2020high,zhu2020identity}.
%
%
The Rank-1/mAP of our method achieves 64.5$\%$/53.6$\%$ and 81.6$\%$/72.1$\%$ on Occluded-Duke and Occluded-REID datasets, which set a new SOTA performance.
Compared to the hand-crafted splitting based method PCB~\cite{sun2018beyond}, our PAT surpasses it by $+21.9\%$
Rank-1 accuracy and $+19.9\%$ mAP on the Occluded-Duke dataset.
This is because our PAT explicitly learns part-aware masks to depress the noisy information
from the occluded regions.
It can be seen that hand-crafted splitting based Re-ID methods and holistic methods with key-points information have similar performance on two occluded datasets.
For example, PCB~\cite{sun2018beyond} and FD-GAN~\cite{ge2018fd} both achieve about $40\%$ Rank-1 score on the Occluded-Duke dataset, indicating that key-points information may not significantly benefit the occluded Re-ID task.
%
%
Compared with PVPM and HOReID, which are SOTA occluded ReID methods with key-points information, our method achieves much better performance,
surpassing them by at least $+9.4\%$ Rank-1 accuracy and $+9.8\%$ mAP on the Occluded-Duke dataset.
%
This is because their performance heavily relies on the accuracy of the off-the-shelf pose estimation models, while our method can capture more accurate human part information in a unified deep model.
%
Furthermore, our PAT also outperforms the methods with the mask learning strategy, including GASM and ISP, which shows the effectiveness of our transformer encoder-decoder architecture and two learning mechanisms.

\noindent\textbf{Results on Partial Datasets.}
\begin{table}[]
\centering
\small
\caption{Performance comparison with state-of-the-arts on Partial-REID and Partial-iLIDS datasets.
Our method achieves the best.}
\label{tab:partial_dataset}
\begin{tabular}{c|cc|cc}
\hline
\multirow{2}{*}{Methods} & \multicolumn{2}{c|}{Partial-REID} & \multicolumn{2}{c}{Partial-iLIDS} \\
\cline{2-5}
                  &Rank-1   &Rank-3   &Rank-1   &Rank-3   \\
\hline
\hline
AMC+SWM~\cite{zheng2015partial}   &37.3   &46.0     &21.0     &32.8     \\
DSR~\cite{he2018deep}       &50.7   &70.0     &58.8     &67.2     \\
SFR~\cite{he2018recognizing}    &56.9   &78.5     &63.9     &74.8     \\
STNReID~\cite{luo2020stnreid}		&66.7	&80.3		&54.6	&71.3		\\
VPM~\cite{sun2019perceive}      &67.7   &81.9     &65.5     &74.8     \\
PGFA~\cite{miao2019pose}      &68.0   &80.0     &69.1     &80.9     \\
AFPB~\cite{zhuo2018occluded}    &78.5   &-        &-        &-        \\
PVPM~\cite{gao2020pose}       &78.3   &87.7        &-        &-        \\
FPR~\cite{he2019foreground}     &81.0   &-        &68.1     &-        \\
HOReID~\cite{wang2020high}      &85.3   &91.0     &72.6     &86.4     \\
\hline
\textbf{PAT}(\textit{Ours})          &\textbf{88.0}    &\textbf{92.3}  &\textbf{76.5}  &\textbf{88.2}  \\
\hline
\end{tabular}
\vspace{-4mm}
\end{table}
To further evaluate our method, we compare the results on Partial-REID and Partial-iLIDS datasets with existing state-of-the-art methods.
Like some previous methods~\cite{sun2019perceive,he2019foreground,gao2020pose,wang2020high}, since the two partial datasets are too small, we train our model on the Market-1501 training set and use two partial datasets as test sets.
Therefore, it is also a cross-domain setting.
As shown in Table~\ref{tab:partial_dataset}, the Rank-1/Rank-3 of our method achieves 88.0$\%$/92.3$\%$ and 76.5$\%$/88.2$\%$ on Partial-REID and Partial-iLIDS datasets, respectively, which outperforms all the previous partial person Re-ID models.
This suggests that the proposed PAT can be solid to address the occlusion problem.
Compared  to the most competing method HOReID~\cite{wang2020high},
%
our PAT significantly surpasses it by $+2.7\%$ Rank-1 accuracy on Partial-REID, while surpasses it by $+3.9\%$ Rank-1 accuracy on Partial-iLIDS,
which demonstrates the effectiveness of our proposed model.

\noindent\textbf{Results on Holistic Re-ID Datasets.}
\begin{table}[]
\centering
\small
\caption{Performance comparison with state-of-the-art methods on Market-1501 and DukeMTMC-reID datasets.
}
\label{tab:supervised_dataset}
\begin{tabular}{c|cc|cc}
\hline
\multirow{2}{*}{Methods}  & \multicolumn{2}{c|}{Market-1501} & \multicolumn{2}{c}{DukeMTMC-reID} \\
\cline{2-5}
                  &Rank-1   &mAP    &Rank-1   &mAP    \\
\hline
\hline
PCB~\cite{sun2018beyond}      &92.3   &77.4   &81.8   &66.1   \\
BOT~\cite{luo2019bag}       &94.1   &85.7   &86.4   &76.4   \\
MGN~\cite{wang2018learning}     &95.7   &86.9   &88.7   &78.4   \\
\hline
VPM~\cite{sun2019perceive}      &93.0   &80.8   &83.6   &72.6   \\
IANet~\cite{hou2019interaction}   &94.4   &83.1   &87.1   &73.4   \\
CASN+PCB~\cite{qian2019leader}    &94.4   &82.8   &87.7   &73.7   \\
CAMA~\cite{yang2019towards}     &94.7   &84.5   &85.8   &72.9   \\
MHN-6~\cite{chen2019mixed}      &95.1   &85.0   &89.1   &77.2   \\
\hline
SPReID~\cite{kalayeh2018human}    &92.5   &81.3   &84.4   &71.0   \\
DSA-reID~\cite{zhang2019densely}  &95.7   &87.6   &86.2   &74.3   \\
$P^{2}$ Net~\cite{guo2019beyond}  &95.2   &85.6   &86.5   &73.1   \\
PGFA~\cite{miao2019pose}      &91.2   &76.8   &82.6   &65.5   \\
HOReID~\cite{wang2020high}      &94.2   &84.9   &86.9   &75.6   \\
FPR~\cite{he2019foreground}     &95.4   &86.6   &88.6   &78.4   \\
\hline
\textbf{PAT}(\textit{Ours})          &95.4		&88.0	&88.8	&78.2  \\
\hline
\end{tabular}
\vspace{-4mm}
\end{table}
We also experiment on holistic person Re-ID datasets including Market-1501 and DukeMTMC-reID.
We compare our method with state-of-the-art approaches of three categories, and the results are shown in Table~\ref{tab:supervised_dataset}.
The methods in the first group are hand-crafted splitting based models. The methods in the second group are attention based approaches. The methods in the third group are extra semantic based methods.
From the results, we can see that the proposed PAT achieves competitive performances with state-of-the-art on both datasets.
Specifically, the Rank-1/mAP of our method achieves 95.4$\%$/88.0$\%$ and 88.8$\%$/78.2$\%$ on Market-1501 and DukeMTMC-reID datasets, respectively.
Our PAT performs better than the hand-crafted splitting based model PCB, because the hand-crafted splitting is too coarse to align the human parts well.
%
Furthermore, the proposed PAT is superior to those approaches with external cues.
Specifically, compared to the pose-guided occluded Re-ID method HOReID~\cite{wang2020high}, our PAT significantly surpasses it by $+3.1\%$ mAP   on Market-1501, while surpasses it by $+2.6\%$ mAP   on DukeMTMC-reID, which shows the effectiveness of the proposed part prototype learning mechanism.
The extra semantic based approaches heavily rely on the external cues for person
alignment, but they cannot always infer the accurate external cues in the case of severe occlusion.
%
The above results also prove that the learnable part prototypes are robust to different views, poses, and occlusions.

\vspace{-1mm}
\subsection{Ablation Studies}
\vspace{-1mm}

In this section, we perform  ablation studies on the Occluded-Duke dataset to analyze each component  of our PAT, including the pixel context based transformer encoder ($\mathcal{P}$), the self-attention layer ($\mathcal{S}$) and the cross-attention layer ($\mathcal{C}$) of the part prototype based transformer decoder and the part diversity mechanism ($\mathcal{D}$).
Note that the part discriminability
mechanism is to make part features maintain identity discriminative, and it is the basis of our model. 
We reomve all the modules and set the ResNet-50 with the average pooling as our baseline, where only a global feature is available.
The results are shown in Table~\ref{tab:ablation}.

\noindent\textbf{Effectiveness of the Transformer Encoder.}
As   shown  in index-2, compared with the baseline model,
when only the encoder is adopted and only the global feature $f^{g}$ is used, the performance is improved by $+7.1\%$  mAP.
This is because the self-attention mechanism of the encoder can capture the pixel context information well.
From index-3 and index-5, we can also see that with the encoder, the performance is improved by $+0.8\%$ mAP since the pixel context aware feature is more robust to background clutters.

\noindent\textbf{Effectiveness of the Transformer Decoder.}
From index-1 and index-3, when the part prototype based decoder is added, the performance is greatly improved by $+12.9\%$ and up to $51.7\%$ mAP.
This shows that the part-aware masks obtained from part prototypes are useful for reducing the influence of background and aligning part features.
%
From index-3 and index-4, when the self-attention layer in the decoder is added,   the performance   is further improved by $+0.9\%$ mAP.
This demonstrates the effectiveness of the local context information propagation among all prototypes.

\noindent\textbf{Effectiveness of the Part Diversity Mechanism.}
From index-5 and index-6, we can see that our full model achieves
the best performance,
which demonstrates the effectiveness of the proposed diversity loss.
By adding the diversity loss, the learnable part prototypes are guided to discover diverse discriminative human parts  for the occluded Re-ID task.

\begin{table}[]
\centering
\small
\caption{Performance comparison with different components.}
\label{tab:ablation}
\begin{tabular}{c|cccc|cccc}
\hline
Index &$\mathcal{P}$    &$\mathcal{S}$ &$\mathcal{C}$      &$\mathcal{D}$  &R-1 &R-5   &R-10  &mAP\\
\hline
\hline
1   &			    &           &           &                     &46.0     &65.7     &71.9     &38.8   \\
2   &$\checkmark$	&           &           &                     &58.2     &75.1     &80.4     &45.9   \\
3   &			    &$\checkmark$    &$\checkmark$    &           &61.9     &76.9     &81.8     &51.7   \\
4   &			    &           &$\checkmark$         &           &59.9     &75.6     &81.2     &50.8   \\
5   &$\checkmark$	&$\checkmark$    &$\checkmark$    &           &63.1     &77.5     &82.1     &52.5   \\
6   &$\checkmark$	&$\checkmark$    &$\checkmark$    &$\checkmark$    &\textbf{64.5}   &\textbf{78.3}     &\textbf{83.4}    &\textbf{53.6}   \\
\hline
\end{tabular}
\vspace{-4mm}
\end{table}
%
\begin{figure}[t]
\includegraphics[width=1.0\linewidth]{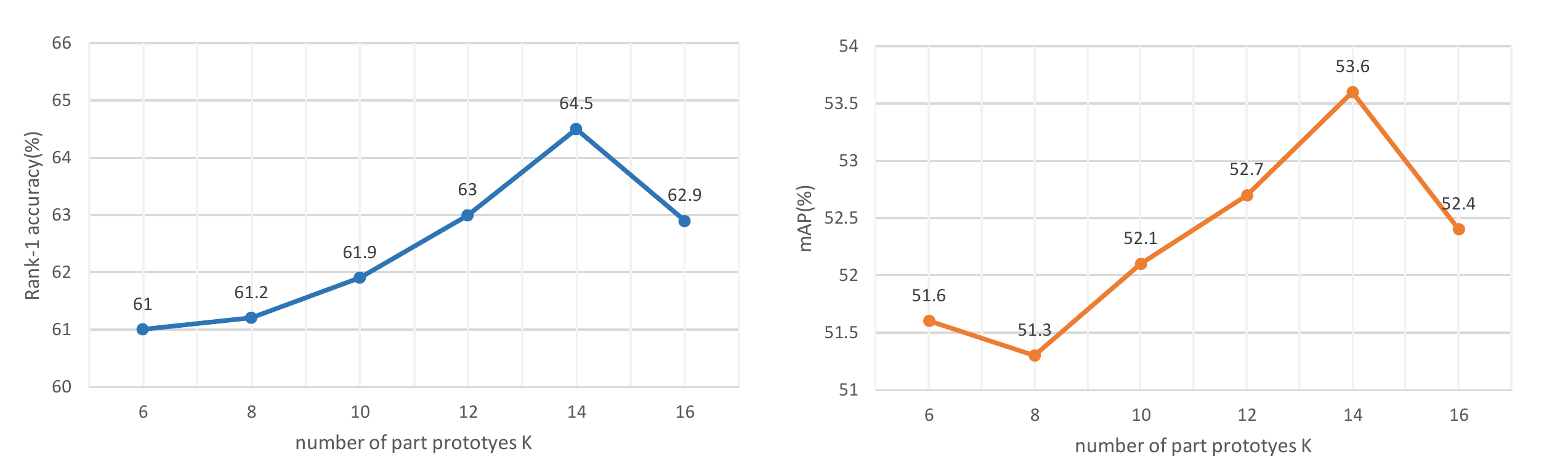}
\caption{Comparison in Rank-1 accuracy and mAP  with different settings of the part prototype number $K$ on Occluded-Duke.
        }
\label{part_num}
\vspace{-3mm}
\end{figure}
\begin{figure}[t]
\includegraphics[width=1.0\linewidth]{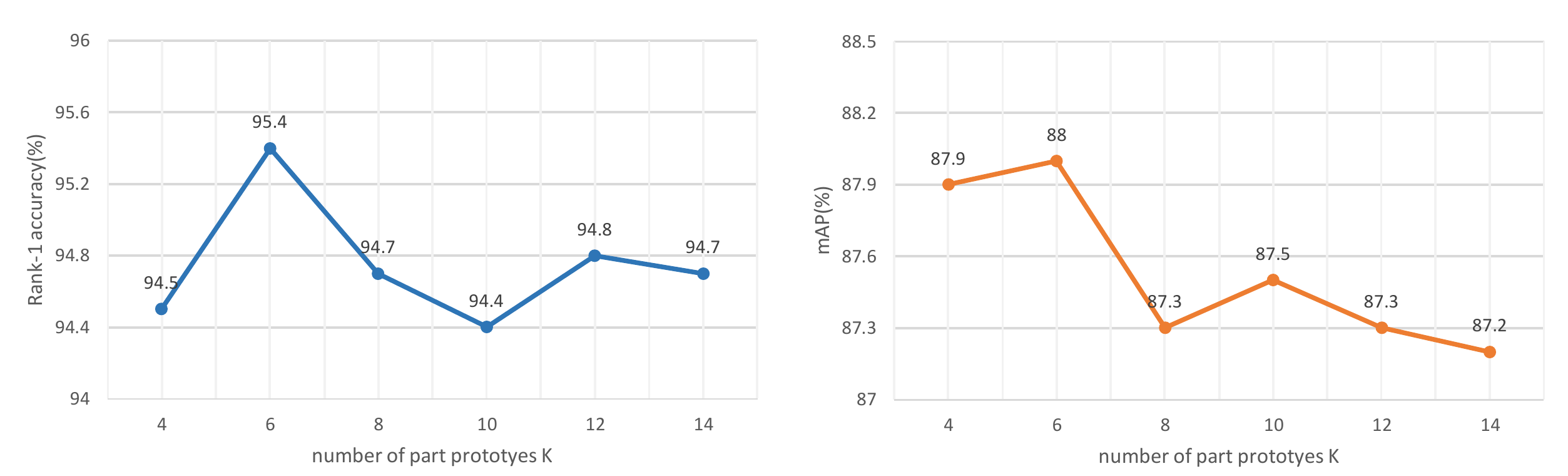}
\caption{Comparison in Rank-1 accuracy and mAP with different settings of the part prototype number $K$ on Market-1501.
        }
\label{part_market}
\vspace{-5mm}
\end{figure}

\noindent\textbf{Analysis of the Number of Part Prototypes.}
The number of part prototypes $K$ determines the granularity of the discovered parts.
We perform quantitative experiments to clearly find the most suitable $K$ on Occluded-Duke and Market-1501 datasets.
As shown in Figure~\ref{part_num},
with $K$ increases, the performance keeps improving before $K$ arrives 14 on Occluded-Duke, while the best performance is achieved when $K$ is set to 6 on Market-1501, as shown in Figure~\ref{part_market}.
We conclude that this is because the scenarios in Occluded-Duke are more complex, and more fine-grained part features would be more useful.
%
%

\vspace{-1.5mm}
\subsection{Visualization of Discovered Parts}
\begin{figure}[t]
\includegraphics[width=1.0\linewidth]{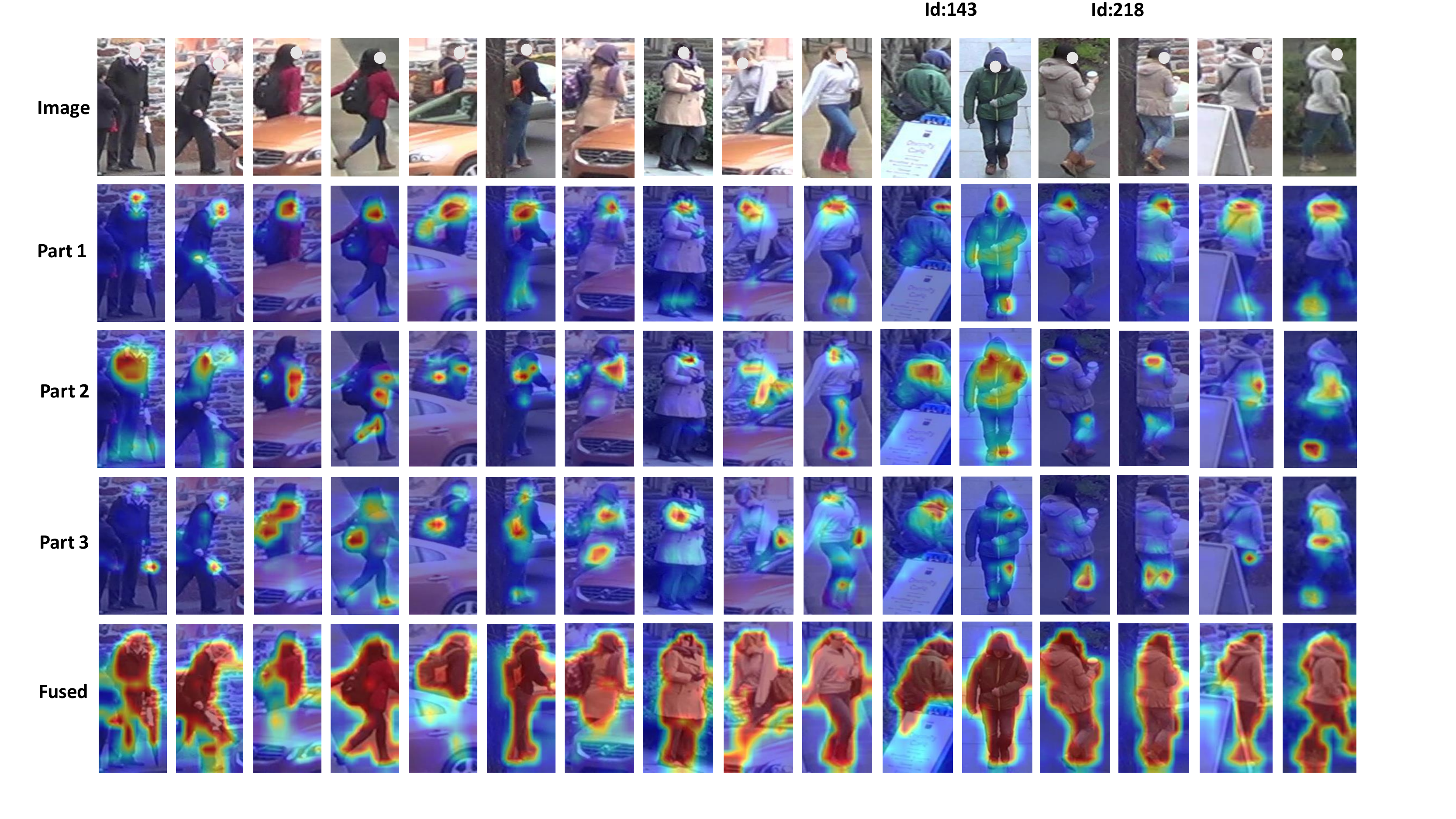}
\caption{
Visualization of the learned part-aware masks.
The final mask is obtained by fusing all the part-aware masks.
As we can see, these part-aware masks mainly focus on different discriminative human parts, including personal belongings.
        }
\label{fig3}
\vspace{-5mm}
\end{figure}

\vspace{-2mm}
We visualize the part-aware masks generated from different part prototypes in Figure~\ref{fig3}.
%
From the results, we can observe that different part-aware masks can successfully capture diverse discriminative human parts for the same input image.
For example, the part-aware mask obtained by the $1^{th}$ prototype mainly focuses on the head region, and the part-aware mask obtained by the $2^{th}$ prototype mainly focuses on the upper body.
This also shows the effectiveness of our proposed part diversity mechanism.
The final mask in Figure~\ref{fig3} is obtained by fusing all part-aware masks together.
We can see that the fused masks almost span  over the whole person
rather than overfit   in some local regions.
In this way, these part-aware masks can reduce the background interference and occlusion, making the network more focus on discriminative human parts for the  occluded Re-ID task.

\vspace{-2mm}
\section{Conclusion}
\vspace{-1.5mm}
In this work, we propose a novel Part-Aware Transformer to discover diverse discriminative human parts with a set of learnable part prototypes for occluded person Re-ID.
%
%
%
To learn part prototypes only with identity labels well, we design
two effective mechanisms, including part diversity and
part discriminability, to discovery human parts in a weakly supervised manner.
Extensive experimental results  for three tasks   on six standard
Re-ID datasets demonstrate the effectiveness of
the proposed method.

\vspace{-2mm}
\section{Acknowledgment}
\vspace{-1.5mm}
This work was partially supported by the National Key Research and Development Program under Grant No. 2018YFB0804204, Strategic Priority Research Program of Chinese Academy of Sciences (No.XDC02050500), 
National Defense Basic Scientific Research Program (JCKY2020903B002),
National Nature Science Foundation of China (Grant 62022078, 62021001, 62071122), Open Project Program of the National Laboratory of Pattern Recognition (NLPR) under Grant 202000019, and Youth Innovation Promotion Association CAS 2018166.

{\small
\bibliographystyle{ieee_fullname}
\bibliography{egbib}
}

\end{document}